%% file: main.tex
\documentclass[runningheads]{llncs}

\usepackage{cite}
\usepackage{amsmath,amssymb,amsfonts}
\usepackage{graphicx}
\usepackage[ruled,vlined,algo2e]{algorithm2e}
\usepackage[table]{xcolor}
\usepackage{comment}
\usepackage{url}
\usepackage{algorithmicx} 
\usepackage{algpseudocode}
\usepackage{caption}
\usepackage{algorithm}
\usepackage{textcomp}
\usepackage{diagbox}
\usepackage{multirow}
\usepackage{xcolor}
\usepackage{breqn}
\usepackage[T1]{fontenc}
\def\BibTeX{{\rm B\kern-.05em{\sc i\kern-.025em b}\kern-.08em
    T\kern-.1667em\lower.7ex\hbox{E}\kern-.125emX}}

\begin{document}


\pagestyle{headings}
\mainmatter
\def\ECCVSubNumber{42}

\title{Behavioural pattern discovery from collections of egocentric photo-streams} 




\titlerunning{Behavioural pattern discovery from collections of egocentric photo-streams}
%
\author{Mart\'in Mench\'on\inst{1}\orcidID{0000-0002-8195-6527} \and
Estefan\'ia Talavera\inst{2}\orcidID{0000-0001-5918-8990} \and
Jos\'e Massa\inst{1}\orcidID{0000-0002-7456-9676} \and
Petia Radeva, Fellow IAPR\inst{3}\orcidID{0000-0003-0047-5172}}
\authorrunning{Mench\'on et al.}

\institute{INTIA, UNCPBA, \textit{CONICET}, Tandil, Argentina \\   \email{\{mmenchon,jmassa\}@exa.unicen.edu.ar}\\  \and
University of Groningen, Groningen, The Netherlands \\  \email{e.talavera.martinez@rug.nl}\\ \and 
University of Barcelona, Barcelona, Spain \\ 
\email{petia.ivanova@ub.edu}
}

\maketitle

\begin{abstract}
The automatic discovery of behaviour is of high importance when aiming to assess and improve the quality of life of people. Egocentric images offer a rich and objective description of the daily life of the camera wearer. 
This work proposes a new method to identify a person's patterns of behaviour from collected egocentric photo-streams. Our model characterizes time-frames based on the context (place, activities and environment objects) that define the images composition. Based on the similarity among the time-frames that describe the collected days for a user, we propose a new unsupervised greedy method to discover the behavioural pattern set based on a novel semantic clustering approach. Moreover, we present a new score metric to evaluate the performance of the proposed algorithm.
We validate our method on 104 days and more than 100k images extracted from 7 users. Results show that behavioural patterns can be discovered to characterize the routine of individuals and consequently their lifestyle.
\keywords{behaviour analysis, pattern discovery, egocentric vision, data mining, lifelogging}
\end{abstract}

\section{Introduction}

The automatic discovery of patterns of behaviour is a challenging task due to the wide range of activities that people perform in their daily life and their diversity, when it comes to behaviour. Advances in human behaviour understanding were made possible, in a great measure, by the widespread development of wearable devices in the last decade, which can be incorporated into clothing, or used as implants or accessories. These devices provide an immeasurable amount of information, which allows  to collect meaningful details about a person's routine. With these sensors it is possible to monitor health \cite{pantelopoulos2009survey}, locate a person at a certain time \cite{sila2016analysis}, the activity carried out at that moment \cite{lara2012survey}, to determine if the person was anxious\cite{huang2016assessing}, and to detect group social interaction \cite{atzmueller2018analyzing}.
However, works that rely on accelerometers or GPS sensors only acquire partial information missing the whole context of the event. The only devices that allow to get visual information about the context and behaviour of the person are wearable cameras. These devices collect egocentric photo-streams that show an objective and holistic view of where the person has been, what activities he/she has been doing, what kind of social interactions he has performed and what objects he has been surrounded, i.e. the camera wearer's behaviour.

\begin{figure}
    \centering
    \resizebox{\columnwidth}{!}{\includegraphics{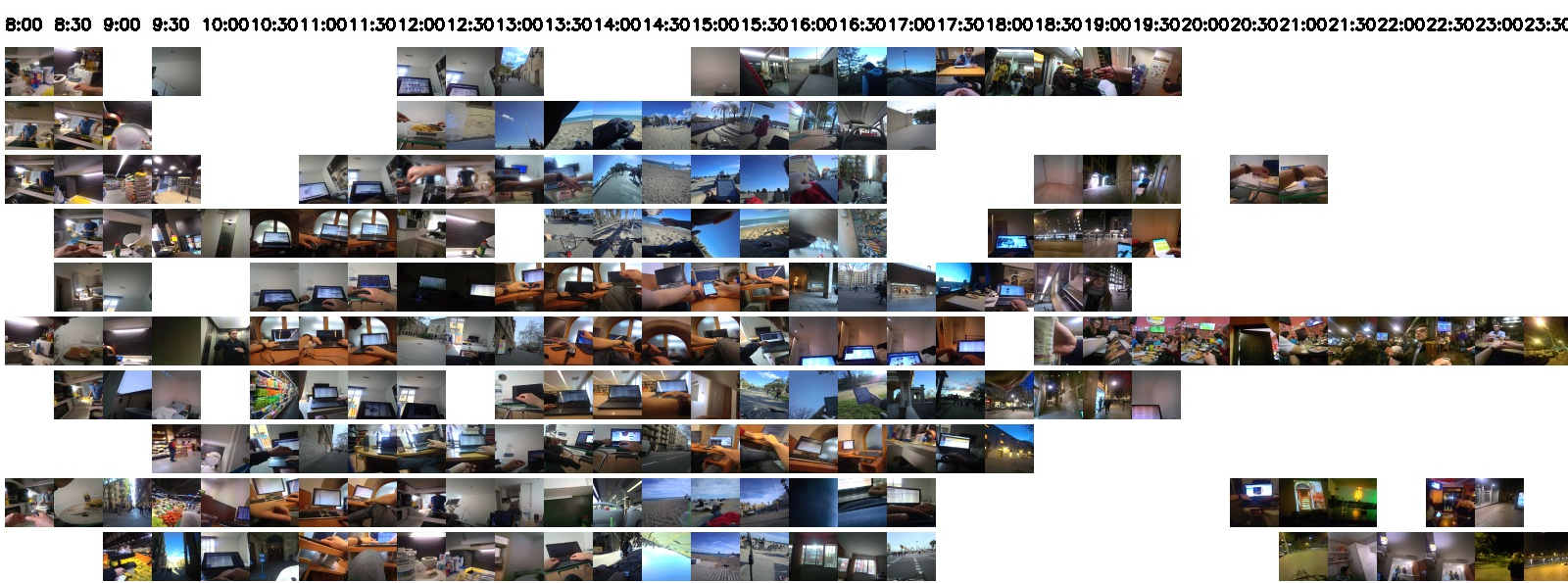}}
    \caption{Visual time-line with collected images by one of the users in the \textit{EgoRoutine} dataset. Rows represent the different collected days and columns indicate time. White spaces correspond to not recorded moments.}
    \label{fig:photosequence}
\end{figure}

Therefore, egocentric photo-streams have shown to be a rich source of information for behaviour analysis \cite{talavera2020topic}. This particular type of images tries to approximate the field of vision of a user. Thanks to the first-person focus of the images, a faithful representation of an individual's day can be obtained. Fig. \ref{fig:photosequence} visually shows a collection of egocentric photo-streams by one of the users. A lot of works has been developed to extract semantic information from egocentric photo-streams like: activity and action recognition \cite{cartas2017recognizing}, places detection \cite{furnari2016recognizing,matei2020deep}, social interaction characterization \cite{aimar2019social,aghaei2018towards}, activities prediction \cite{damen2018scaling}, routine days discovery \cite{talavera2019unsupervised,talavera2020topic}, etc. However, up to our knowledge there is no work on  automatic detection of routine patterns in collection of egocentric photo-streams.

In this work, we propose a new method for  discovery of sets of patterns describing someone's routine from collections of egocentric photo-streams. Given a collection of images, our model is able to describe the collected days with time-frames characterized by recognized objects, activities, and location. Afterwords, a new greedy clustering algorithm is defined to detect patterns of time-frames with similar context based on computed semantic similarity between time-frames labels. Hence, patterns are presented by time-slots with high semantic similarity, and can be used to describe behavioural habits.

The contributions of this work are three-fold:
\begin{itemize}
    \item To the best of our knowledge, this is the first work that addresses the automatic discovery of patterns of behaviour from collections of egocentric photo-streams.
    \item A novel greedy clustering algorithm is proposed to efficiently find patterns in collections of time-lines composed by similar concepts describing time-slots. Special attention is paid to redefine the distance between time-slots mixing categorical and numerical information.
    \item A new score measure is defined to validate extracted patterns from egocentric images.
\end{itemize}

The rest of the paper is organized as follows: in Section \ref{section_SOA} we describe related works. In Section \ref{section_methodology}, we describe our proposed pipeline for behavioural pattern discovery from egocentric photo-streams. In Section \ref{section_experiments}, we describe the performed experiments and obtained results. Finally, we summarize our findings and future research lines in Section \ref{section_Conclusiones}.

\input{2_stateofart}

\input{3_methodology}

\input{4_experiments}

\section{Conclusions}
\label{section_Conclusiones}
In this work, we  presented a novel pattern discovery model for the automatic detection of behavioural habits from collections of visual logs collected by wearable cameras.
The validation of the obtained patterns indicates that our model is valuable and robust, and sets the basis for further research in the field of behaviour understanding from egocentric images.
Our proposed model can be applied to other tasks for pattern discovery since it evaluates the similarity among samples of different time-sequences. Further research will explore its applicability to other tasks.

Results show that the chosen descriptors allow us to discover patterns. However, more analysis is required to fully understand the relevance of the different subsets of descriptors ('activity', 'scene', and 'objects'). For instance, objects with a high recognition accuracy are considered but  frequency of appearance of these objects within the event is not considered. We could use tools such as Word2Vec \cite{DBLP:journals/corr/abs-1301-3781} to evaluate the semantic similarity of recognized objects. These are questions that we plan to continue evaluating to reach a wider overview about the relevance of the different descriptors. On the other hand, we are interested in discovering more complex patterns composed of combination of events (e.g. the individual used to walk on the street after work). We consider that this work is only the beginning of the research line to discover meaningful patterns of human behaviour from egocentric photo-streams. 

\section*{Acknowledgment}
This work was partially founded by projects RTI2018-095232-B-C2, SGR 1742, CERCA, Nestore Horizon2020 SC1-PM-15-2017 (n° 769643), and Validithi EIT Health Program. The founders had no role in the study design, data collection, analysis, and preparation of the manuscript.

\bibliographystyle{IEEEtran}
\bibliography{bibliography}

\end{document}

%% file: 2_stateofart.tex
\section{Related works}
\label{section_SOA}

Pattern discovery is a well-studied topic in computer vision, covering from molecular pattern discovery \cite{brunet2004metagenes}, to bird population analysis \cite{hassell1976patterns}, passing through text mining or speech analysis applications, such as \cite{zhong2010effective} and \cite{park2007unsupervised}, respectively. 

The analysis of behavioural patterns is a relevant topic when studying an ecosystem. However, describing the behaviour of human beings is a difficult task due to the high diversity among individuals, as described in \cite{benchetrit2000breathing} as well as lack of common devices and tools to acquire rich and complete information about individual's behaviour. Human behaviour was studied employing several strategies that use different sources of data, as described in the survey in \cite{borges2013video}. However, most of the approaches propose supervised learning, using techniques to build models that learn from predefined activities. In contrast, proposing models to discover patterns in an unsupervised manner from the collected data, without restricting the procedure to a limited and predefined set of activities,  still remains an open question.

In \cite{kim2009human}, the authors describe the discovery of patterns as information extracted ``directly from low-level sensor data without any predefined models or assumptions''. They proposed a tool for activity classification, where activities are represented as a pattern of values from the collected sensors. They focused on the classification of simple predefined human activities. One of the shortcomings is that there is only a limited and pre-defined set of activities that not always describe the variety of human lifestyle. Moreover, in many situations specially using wearable cameras there is no information about the activity of the person. This fact could be compensated by information about the place and the context i.e. surrounding objects appearing in the egocentric photo-streams. 

Thematic pattern discovery in videos was addressed in \cite{zhao2011discovering}. The authors proposed a sub-graph mining problem and computed its solution by solving the binary quadratic programming problem. However, this method is not capable of seeking for common patterns in several time-sequences. Moreover, someone's behaviour is not an one-theme topic, but a non-defined task where an efficient algorithm for multiple pattern discovery is needed.

Related to egocentric vision, behaviour analysis from egocentric photo-streams has been previously addressed in \cite{talavera2020topic}. The authors proposed to use topic modelling for the discovery of common occurring topics in the daily life of users of wearable cameras. Their  aim was of discovering routine days in individual's diaries. However, their work is not capable of automatically finding patterns as part of the day activities, and what they represent. In this work, instead of aiming to describe what a routine day is, we claim to discover what are the patterns (e.g. breakfast at 8am, office at 10 am) that compose the lifestyle of the camera wearer, in an unsupervised manner.

%% file: 3_methodology.tex
\section{Behavioural pattern discovery}
In this section, we describe our method for pattern discovery from egocentric photo-streams.

\label{section_methodology}

\subsection{Days characterization by semantic features extraction}
Many empirical studies show that most people lead low-entropy lives \cite{Eagle2006}, following certain regular daily routines over long time, although such routines vary from one person to another. Hence, it can be meaningful to develop automatic methods to capture the regular structures in users’ daily lives
and make use of that information for user classification or behavior prediction.
People used to perform similar events in their routine days like working at office in the morning, having lunch at 1pm, etc. Here, we define a pattern as a sequence of elements regarding human behaviour (activities, place, environment represented by its objects around) which descriptors are shared and recurrently occur throughout time. In our case, elements are extracted from the egocentric images that compose photo-streams.
In order to obtain a person routine pattern, we propose a novel unsupervised learning approach focusing on the three important factors that are able to describe what a user is doing at any certain moment: the scene, the activity that he/she is  carrying out and his/her context expressed by the objects present around him/her:

\begin{description}

\item [Activity detection:]
For characterization of the activities we use the network proposed in \cite{cartas2017recognizing} to classify a given image as belonging to one of 21 activities of Daily Living.
\item[Object detection:]
We use the Yolo3 Convolutional Neural Network (CNN) model \cite{yolov3} to extract the appearing objects in the images.
\item [Scene detection:]
For classifying the scene in the single images from the photo-stream, we applied the \textit{VGG16-Places365} \cite{zhou2017places} CNN, trained with the ``Places-365'' dataset consisting of 365 classes.

Note that both object detection and scene detection are not fine-tuned to the egocentric photo-streams that probably leads to suboptimal performance on label detection. This fact is compensated by defining a method for patterns discovery tolerant (as it will be made clear in the Validation section) to suboptimal labels extraction.
\end{description}

\subsection{Graph representation of the day}

\begin{figure}[h!]
    \centering
    \includegraphics[width=0.65\columnwidth]{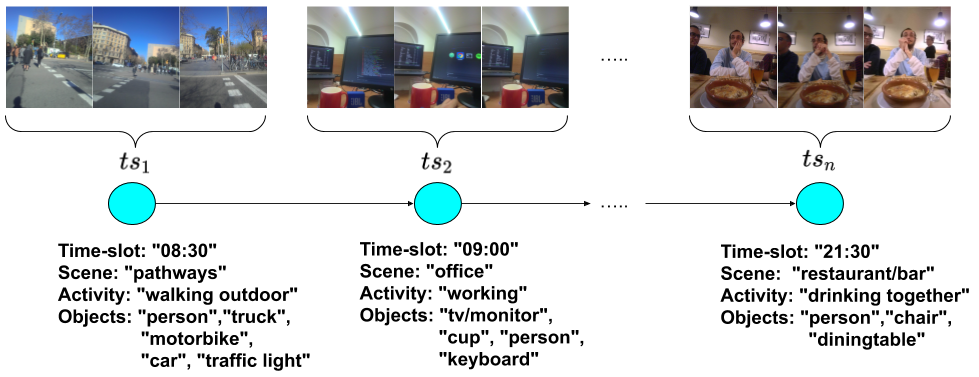}
    \caption{Illustration of created time-slots for sequences of images corresponding to 30' within the images time series. We indicate the automatic extracted and selected labels to describe the time-slot.}
    \label{fig:exampletimeslots}
\end{figure}

We divide the user's days into half-hour time-slots each one.  Each of the time-slots is considered a node in a linear graph  (see Fig. \ref{fig:exampletimeslots}). We characterize every node by evaluating the detected concepts in the images that compose them. We analyze the images by using pre-trained networks to recognize the scene, activity and objects in the scene that they depict. At a node level, we select the scene and the activity label with highest occurrence as the ones describing the time-slot. We also add as descriptor the objects that appeared in several images (in our case, in more than ten) in order to ensure that they are representative enough of that time-frame. 
Fig. \ref{fig:exampletimeslots} illustrates how time-slots are formed and their descriptors at labels level.

\subsection{Detecting Similar Time-slots}
\begin{figure} 
    \centering
    \includegraphics[width=0.7\columnwidth]{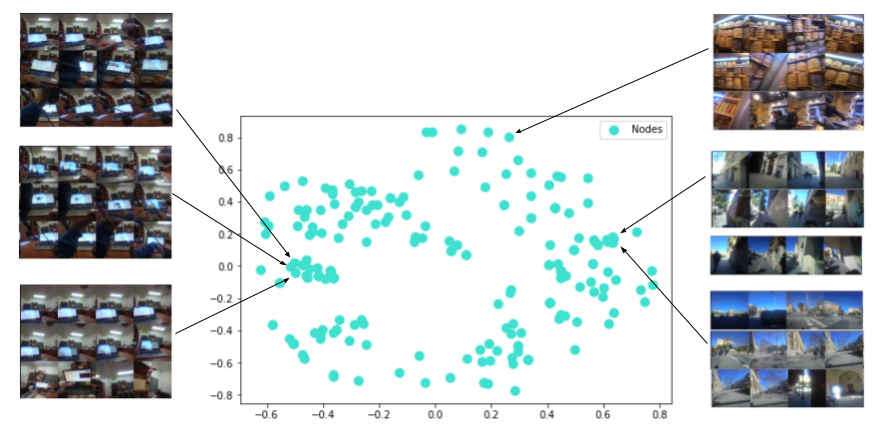}
    \caption{Output of the multi-dimensional scaling. The input is the distance matrix among nodes computed by the defined distance in Eq. (\ref{distance_equationo}). Data samples correspond to nodes, i.e. time-slots in the collection of egocentric photo-sequences.}
    \label{fig:mds}
\end{figure}

Once linear graphs are obtained, we define a metric to compare them. In particular we compare nodes corresponding to the same time-slot assuming that time should be an inherent feature of patterns (e.g. eating at different time means having breakfast, lunch or dinner). The distance between nodes is represented by:
\begin{dmath}
d((S_1,A_1,O_{11},O_{12},..O_{1n1}),(S_2,A_2,O_{21},O_{22},..,O_{2n2})) = D(S_1$<>$S_2)+D(A_1$<>$A_2)+DJ(O_1,O_2)
\label{distance_equationo}
\end{dmath}
where $S_i, i=1,2$ is the scene, $A_i$ is the activity and $O_{i,j}, \; j=1,...n_i$  are the objects of the i-th node. $D(S_1<>S_2)$ is the binary distance between nodes scenes, $D(A_1<>A_2)$ is the binary distance between nodes activities and $DJ(O_1 , O_2)$ is the Jaccard distance \cite{real1996probabilistic} between nodes objects. Once obtained the distances between all pair of nodes for the same time-slot,  we apply multi-dimensional scaling (MDS) \cite{borg2005modern} to arrange  all the nodes in a two-dimensional space. In this way, nodes that were very different from each other are sent far apart and very similar nodes from each other are grouped together. In Fig. \ref{fig:mds}, the nodes spatial distribution after the application of the MDS is shown. Note that since nodes are represented by their labels, they do not have a numerical representation to visualize in space. However, applying the distance defined in eq. (\ref{distance_equationo}) and the MDS allow us to  observe the spatial relation between the nodes and to visually identify groups of highly similar images.


\subsection{Aggregating similar time-slots}

To find the patterns, an heuristic aggregating greedy algorithm is proposed. We start with a seed that is settled from the two nodes 
\begin{equation*}
  n_1=n(i_1,j_1), n_2=n(i_2,j_1)  
\end{equation*}
that are at a minimum distance being from the same time-slot: 
\begin{equation*}
  d(n(i_1,j_1),n(i_2,j_1)) = min_{k,l,m}(d(n(i_k,j_l),n(i_m,j_l)). 
\end{equation*}
The idea behind the algorithm is to find similar nodes considering as neighbours contiguous time-slots and/or different days from the same user. By this way, the algorithm looks for a candidate pattern consisting in recurrent similar activities.

A cluster agglomeration of these nodes is carried out where the nodes from the same time-slot or neighbour time-slots are agglomerated using the variance as a criterion to decide which neighbour time-slot to merge. 
Let us consider that the time-slots aggregated to the pattern $P$ are $P=\{n_1,n_2\}$. And $S$ is the rest of the nodes. From all neighbour nodes $n_k$, we select the node that added to $P$ contributes causing the smallest variance increase. In order to compute the variance we applied the position of the node provided by the multi-dimensional scaling. See Algorithm \ref{algorithm} for the formalization of our proposed method.

\begin{algorithm}[H]
\SetAlgoLined
initialization: characterization of nodes (n) with the images' detected concepts$\;$
\begin{itemize}
    \item Let $n_0=n(i_0,j_0)$ and $n_1=n(i_1,j_0)$ are the closest nodes so that $d(n(i_0,j_0),n(i_1,j_0)) = min_{k,l,m}(d(n(i_k,j_l),n(i_m,j_l))$
    \item Let $v_1 = d(n_0,n_1)$
    \item $S_0=\{n_i\}$ is the set of all nodes 
    \item $P_1=\{n_0,n_1\}$, $S_1=S_0 \setminus \{n_0,n_1\}$, $t=1$
    \item Until $S_t <> empty$ set or $v_t>K$, do:
    \item Let $n_t =argmin_{n_l \in neighbours(S_{t-1})}  var(P_{t-1}+n_l)$ \# now we evaluate all neighbours from the aggregated sets 
    \item $P_t=P_{t-1} \bigcup \{n_t\}$, $S_t=S_{t-1} \setminus \{n_t\}$, $t=t+1$
\end{itemize}  
 \caption{Pattern discovery from collections of egocentric photo-streams}
 \label{algorithm}
\end{algorithm}

The computational complexity of this algorithm is $O(ds*ts*N)$ where $ds$ is the number of days to analyze, $ts$ is the amount of time-slots of the day, and $N$ is the number of nodes.

\subsection{Finding patterns}
Let us consider that the nodes aggregated are $(n_1,n_2,....n_M)$ and the variance evolution is $v=(v_1,v_2,...v_M)$ adding all neighbour nodes.
We are interested in extracting a cluster of semantically similar neighbour nodes (e.g. time-slots). We apply the assumption that similar nodes would have small variance while adding dissimilar nodes would lead abrupt increase in the variance. Note that by the aggregation process the variance evolution is monotonically increasing function. Abrupt change can be detected by looking for the maximum of the first derivative of $v$ and zero-crossing of the second derivative. 
So to find the cluster with similar nodes, we  smooth $v$ through a Gaussian filter and then apply the first and second derivative to detect the separation between clusters. We find  $n_k$ so that there is a zero-crossing for the second derivative $$n_k: d^2(v_k)/dt^2=0$$ and the first derivative has a  value over a given threshold $T$  $$d(v_k)/dt>T$$ we consider that $(n_1,n_2,...n_{k-1})$ form a cluster. In practice, there could be several candidates that fulfil the conditions for the first and second derivatives. In order to choose the optimal one, we define a new criterion for pattern discovery. On one hand, we are interested in selecting patterns that form separated clusters maximizing inter-class distance and minimizing intra-class distance. A good measure for this is given by the  silhouette measure of the clustering. The Silhouette score is described by equation (\ref{eq:silh}), where  $a(i)$ is the average distance between point $i$ and points withing the same cluster, and $b(i)$ is the minimum average distance from $i$ to points of the other clusters: 
\begin{equation}
sl = silhouette_{score} = \frac{b(i) - a(i)}{max(a(i), b(i))}\label{eq:silh}.
\end{equation}

On the other hand, we are interested in obtaining routine patterns that should be recurrent that is being spread on as much as possible days and  they should cover significant time. To this purpose, we introduce the scoring function shown in equation (\ref{eq:scorf}) to evaluate the goodness of the detected pattern:
\begin{equation}
sc(P_{k-1})=sl(P_{k-1})+t_{rpr}
\label{eq:scorf}
\end{equation}
where $t_{rpr}$ expressed by equation (\ref{eq:Representativeness}) is measuring the representativity of the pattern. To this purpose, this term measures the number of days the pattern covers (the higher the better) and the average time of prolongation $\hat{l}$ of the pattern during the days: 

\begin{dmath}
t_{rpr}= \frac{1}{\#patterns} \sum_{i=1}^{\#patterns}
(\# ds_{i} / \#ds +
\#I_{i}/ max(\#I(1hour), (tm(I_{i,n_i})-tm(I_{i,1}))*frq)
\label{eq:Representativeness}.
\end{dmath}
where $ds$ stands for days, $I$ stands for image, $tm$ stands for time and $frq$ stands for camera frequency resolution (in our case, $frq = 0.5$ fpm).

\begin{figure}[ht]
    \centering
    \includegraphics[width=0.7\columnwidth]{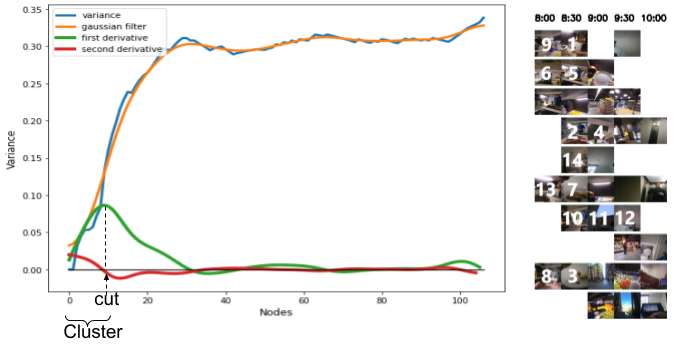}
    \caption{Example of a fluctuation of the cluster variance as time-slots are added (blue). Filtered variance signal with a Gaussian filter (orange). First and second derivative (green and red, respectively). Cluster cut with the maximum value of the second derivative. On the right, visual representation of the sequentially added time-slots to one identified cluster, i.e. a discovered pattern.}
    \label{fig:Variance}
\end{figure}

Once a pattern is detected, those time-slots are discarded and the rest of time-slots are processed using the same procedure in order to find other patterns of behaviour. Fig. \ref{fig:Variance} visually describes the process of finding a pattern.

%% file: 4_experiments.tex
\section{Experiments}
\label{section_experiments}
In this section, we present the dataset used, the validation metrics, the experimental setup and the final results.

\subsection{Dataset}

In this work, we employ the \textit{EgoRoutine} dataset proposed in \cite{talavera2020topic}. The \textit{EgoRoutine} dataset consists of unlabelled egocentric data of 115430 images collected on 104 days captured by 7 distinct users. The recorded images capture the daily lives of the users. In Fig. \ref{fig:photosequence} we visually present a subsampled mosaic of  images collected by one of the users. Table~\ref{tab:routine_users} presents an overview of the dataset distribution.

\begin{table}[h!]
\vspace{-1em}
    \centering
    \caption{Distribution of the data in the  \textit{EgoRoutine} dataset.}
    \resizebox{0.7\columnwidth}{!}{
    \begin{tabular}{c|c|c|c|c|c|c|c|c}
         & User 1& User 2 & User 3 & User 4 & User 5& User 6& User 7 & Total \\
         \hline \hline
         \#Days & 14 & 10 & 16 & 21 & 13 & 18 & 13 & 104 \\
         \#Images & 20521 & 9583 & 21606 & 19152 & 17046 & 16592 &  10957 & 115430 \\
    \end{tabular}
    }
    \label{tab:routine_users}
\end{table}

\subsection{Validation metrics}

The proposed model is applied following a personalized approach since patterns are highly individual, i.e. we evaluate the performance at user level by evaluating over collected days by each user. The model applies unsupervised learning thus, no labels nor dataset split are required. 
Since our method discovers the patterns in an unsupervised manner, a proper performance should be 
evaluated quantitatively and qualitatively. 



We quantify the quality of the found patterns with the Silhouette score \cite{rousseeuw1987silhouettes} and the temporal spread of the patterns. The Silhouette metrics describes the relatedness of each point w.r.t. the cluster group it has been assigned to. 

Qualitative measures usually come down to human judgement. To this end, we created a survey that showed to 13 different individuals: (1) the set of images that visualize the collected days of the users, and (2) the found patterns. Individuals were asked the following four questions:

\begin{enumerate}
    \item Given the following collected egocentric photo-streams, reflecting about the life of the camera wearer, can you see patterns of behaviour?
    What patterns do you see in this mosaic of images showing the days of user \#ID?
    \textit{Note that "a pattern is defined as a concurrent habit of the camera wearer".}
    \item Given these two discovered sets of patterns, which pattern set do you think better represents the user's behaviour?.
    \item Given the previous recorded days for user \#ID, do you think the found patterns adequately represent the user's behaviour?."
    \item Do you consider that the found patterns are predominantly: a) sub-patterns, b) no patterns, c) merged patterns, or d) correct patterns?
    \end{enumerate}

\subsection{Experimental setup}

In this work, we use several available models for the extraction of descriptors from the images with the aim of characterizing the user's days. 
\begin{itemize}
    \item We rely on Places365 \cite{zhou2017places} to recognize the place depicted in the images. We use the top-1 label to represent the scene.
    \item Objects are detected in the frames using the Yolo3 \cite{yolov3} CNN model. We keep only the detected objects with a given class probability $> 0.5$.
    \item The network introduced in  \cite{cartas2017recognizing} is used to identify the activity that is described in the frame. We use the top-1 label to represent the scene.
\end{itemize}

\subsection{Results}

In this subsection, we evaluate our proposed method for pattern discovery.

\begin{figure}[ht!]
    \centering
    \includegraphics[width=\columnwidth]{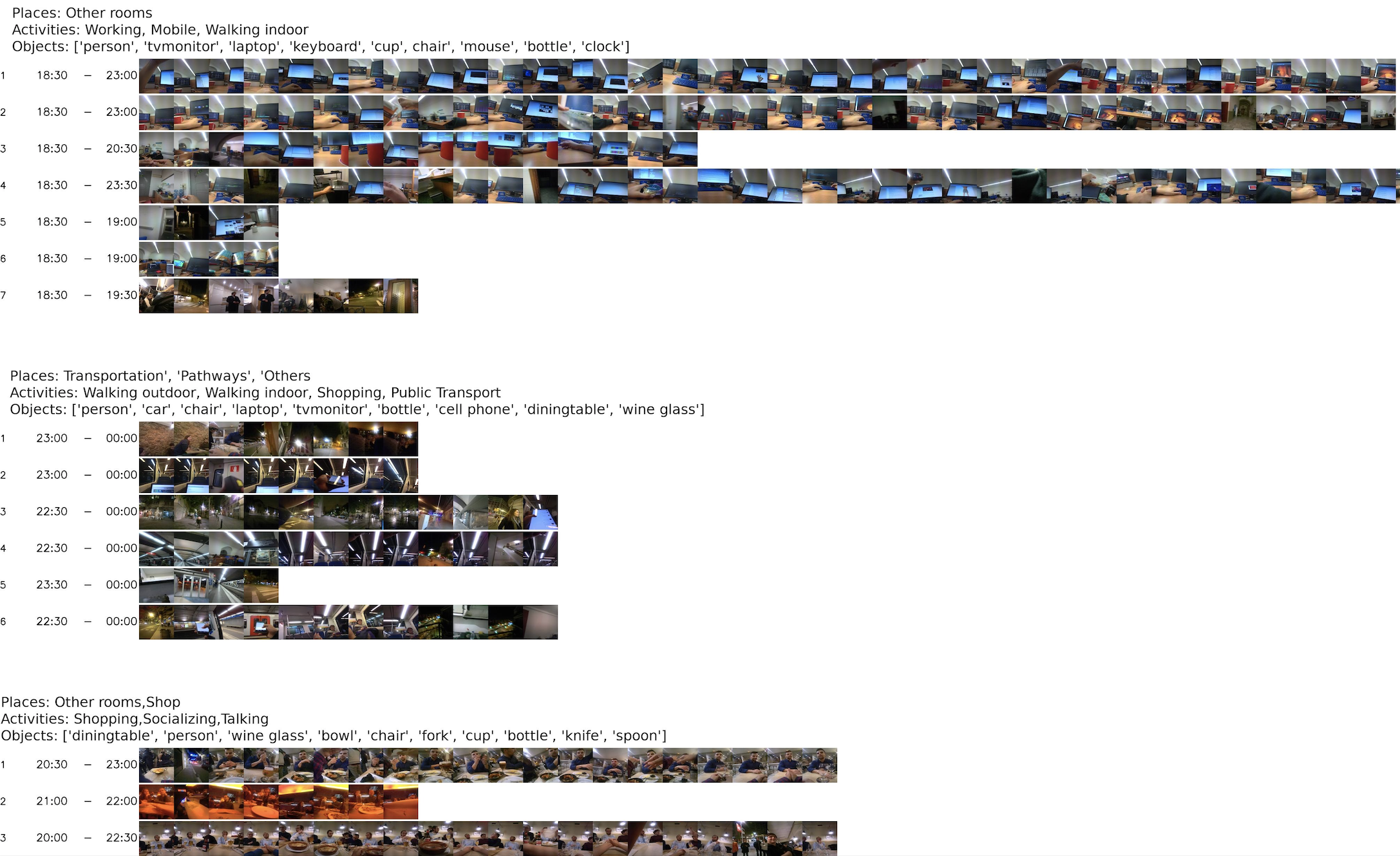}
    \caption{Some discovered patterns for User 1 and the labels used to analyze them.}
    \label{fig:discoveredpatterns_labels}
\end{figure}

In Fig. \ref{fig:discoveredpatterns_labels} we show some samples of discovered patterns for User 1. Each pattern is described by the places, activities, and objects recognized in the images that compose it. This allows us to get a better understanding of what the pattern represents. Given a pattern, each row indicated the number of times it appears in the collection of photo-streams. The time-span of such event is indicated on the left side of the pattern. We can observe how the first pattern describes an office-related activity, while the second and third pattern describe a commuting- and social eating-related activities, respectively. 

\begin{figure}[h!]
    \centering
    \includegraphics[width=0.8\columnwidth]{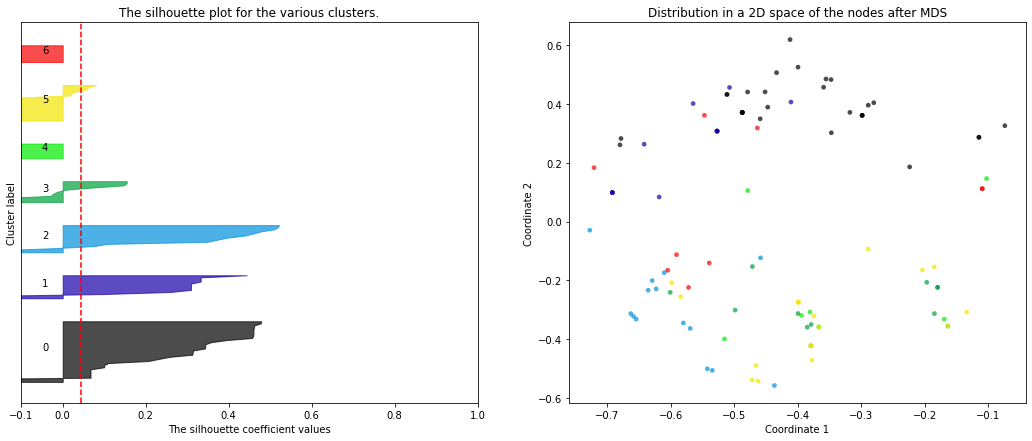}
    \caption{(Left) Distribution in a 2D space of the nodes after MDS. Colours indicate the cluster label. (Right) Clusters quality measure with the Silhouette Score for the computed clusters for User 2.}
    \label{fig:Silhouette}
    \vspace{-0.4em}
\end{figure}

We rely on the Silhouette score metric to evaluate the quality of the found cluster of images that are finally considered as a pattern. In Fig. \ref{fig:Silhouette}, we present an example of clusters silhouettes. On the right figure we can see the Silhouette coefficient values for the different found clusters. High values of Silhouette score indicate high intra-class and low inter-class similarity among samples within the different clusters. On the right, we have a 2D visualization of the nodes under analysis. The coordinates are obtained from the computed distance matrix with multi-dimensional scaling method. Colours indicate the clusters that are assigned to the nodes. We can observe that even though some elements overlap, clusters can be visually identified.

\begin{figure}[h!]
    \centering
    \includegraphics[width=0.7\columnwidth,height=0.3\columnwidth]{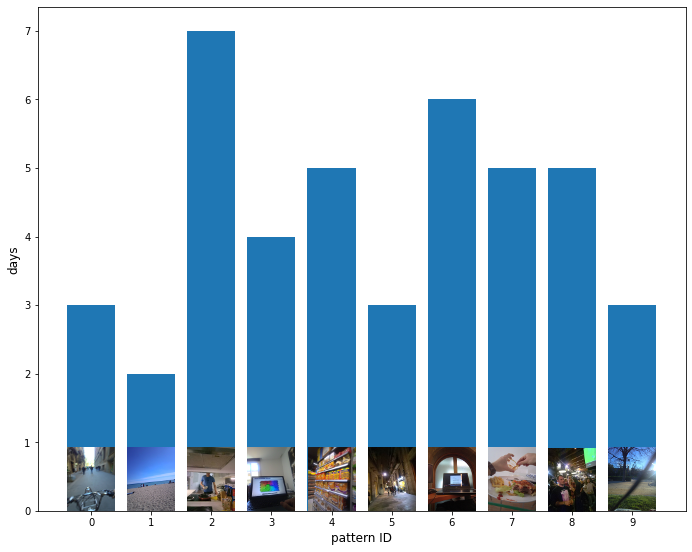}
    \caption{Histogram of discovered patterns occurrence. X-axis indicate the pattern ID - we add a sample image as an illustration of the scene that the pattern represents. Y-axis indicate the number of days in which the pattern appears.}
    \label{fig:histogram_patterns}
\end{figure}

Furthermore, we evaluate the occurrence of patterns in the daily life of the camera wearer. A good estimation is to visually evaluate with a histogram the discovered patterns. Fig. \ref{fig:histogram_patterns} indicates the occurrence of patterns for User 2. Samples images have been added to give visual information of what the pattern depicts. We solely indicate the number of days in which the pattern appears. We can observe how the user goes to the supermarket 5 times (pattern 4), bikes 3 days (pattern 0), or works with the computer 10 days (patterns 3 and 6).

\begin{figure}[h!]
    \centering
    \includegraphics[width=\columnwidth]{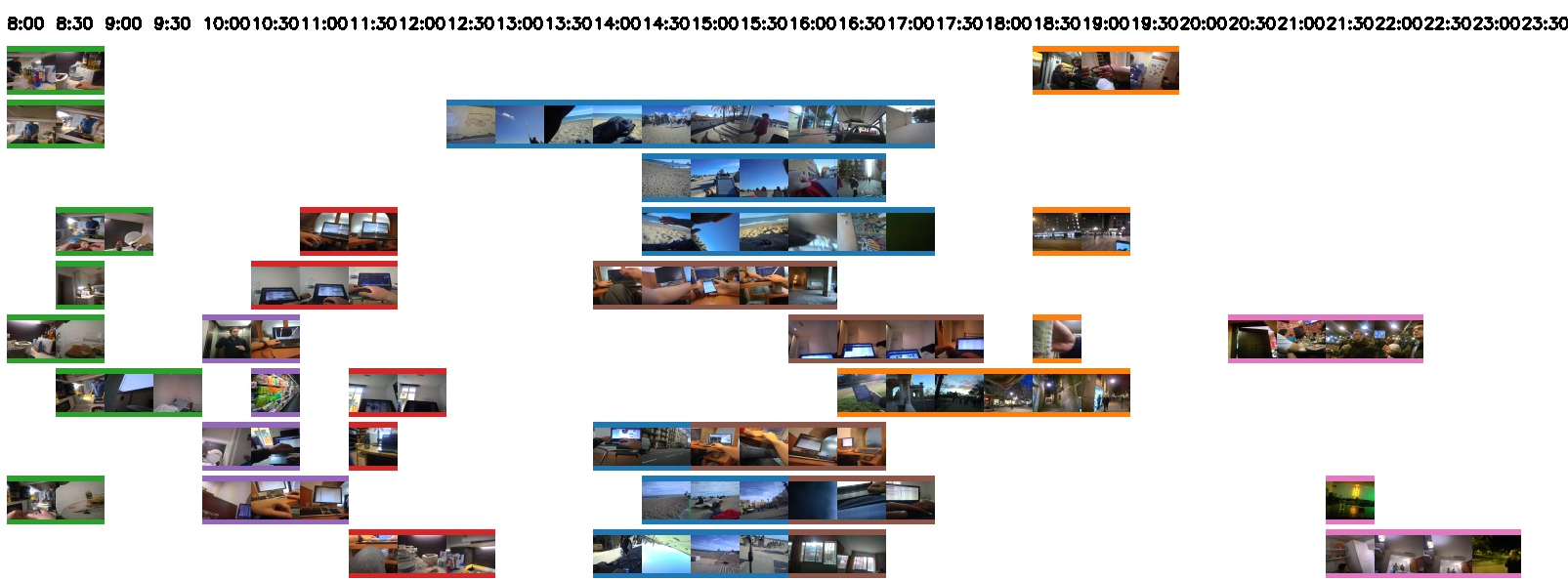}
    \caption{Discovered patterns given a user's collection of photo-streams. Rows represent collection of days. Columns represent time-slots of 30'. One sample image is shown per time-slot. Colours indicate different discovered patterns.}
    \label{fig:user_02_patterns}
\end{figure}

In Fig. \ref{fig:user_02_patterns}, we can observe the discovered patterns for User 2. We use different colours to represent different patterns. The time-slots that are not associated to any pattern are not included.  We can see how the randomly placed seed spread throughout time-slots that are visually alike, i.e. define similar scenes. For instance, the user has the habit of going to the beach at lunch time (blue colour). The user also works in the morning (red colour) with a computer. We can observe how some visually similar time-slots composed of the same activity belong to different patterns. After visually inspecting these cases, we observed that this is due to the fact that different objects appeared in these scenes. As an example of this, the purple and red patterns found from 10:00h to 12:00h, indicate that the user was working in front of the PC while interacting with different objects.

\begin{figure}[h!]
    \centering
    \includegraphics[width=\columnwidth]{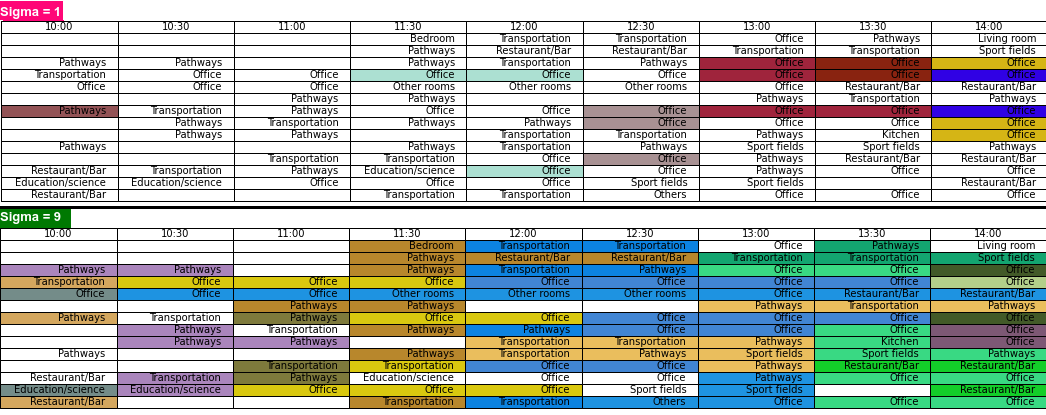}
    \caption{Found User 1's patterns shown with different colours, applying Gaussian smoothing with Sigma=1 and Sigma=9}
    \label{fig:sigma}
\end{figure}

We evaluated the influence of the parameter sigma of the Gaussian applied to the variance evolution curve in order to assess the first and second derivatives. In Fig. \ref{fig:sigma} two set of patterns can be shown that correspond to sigma 1 and sigma 9 respectively. It can be seen that with sigma 1 a greater number of small patterns are found, while with sigma 9 long-term patterns (in times-lots and days) are observed that are more consistent.

A cluster varies its inner variance when including new elements. We use the score defined in Eq. (\ref{eq:scorf}) to assess the quality of the formed clusters with different threshold values $T$ on the first derivative. We tested threshold values within the range [0 - 0.05]. In Table \ref{tab:selectedthreholds}, we present as quantitative results the automatically selected thresholds on the first derivative for every user that yields the maximum score.

\begin{table}[ht]

\centering
\caption{Automatically selected thresholds for each user}
\begin{tabular}{|c|c|c|c|c|c|c|c|}
&  User 1 & User 2 & User 3 & User 4 & User 5 & User 6 & User 7 \\  \hline
Threshold & 0.002  & 0.006  & 0.008  & 0.01  & 0.006  & 0.01  & 0.006  \\
\end{tabular}
\label{tab:selectedthreholds}
\vspace{-2em}
\end{table}

\begin{figure}[h!]
  \centering
  \begin{minipage}[b]{0.45\textwidth}
    \includegraphics[width=\textwidth]{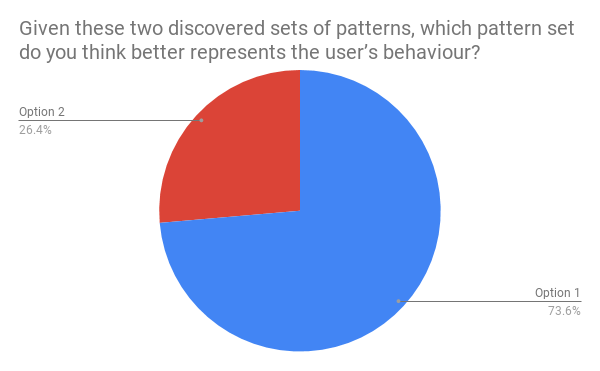}
    \caption{Answers to Question 2.}
    \label{Question2}
  \end{minipage}
  \hfill
  \begin{minipage}[b]{0.45\textwidth}
    \includegraphics[width=\textwidth]{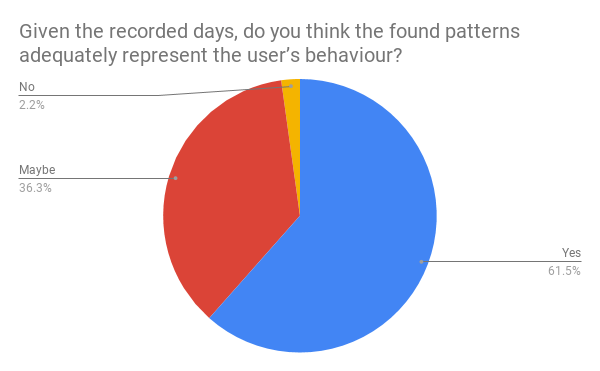}
    \caption{Answers to Question 3.}
    \label{Question3}
  \end{minipage}
  \centering
  \begin{minipage}[b]{0.45\textwidth}
    \includegraphics[width=\textwidth]{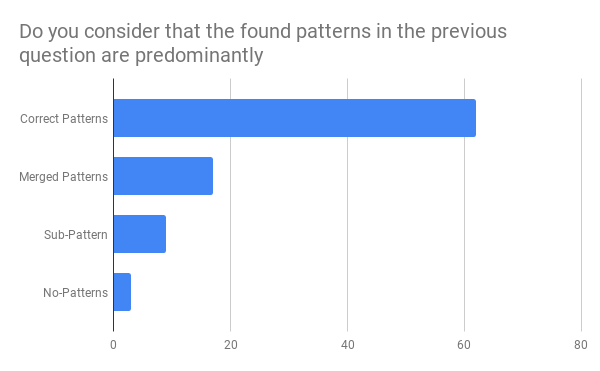}
    \caption{Answers to Question 4.}
    \label{Question4r}
  \end{minipage}
  \hfill
  \begin{minipage}[b]{0.45\textwidth}
    \includegraphics[width=\textwidth]{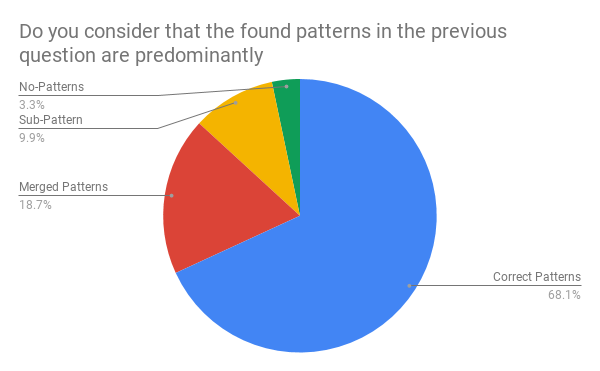}
    \caption{Answers to Question 4 - Percentages.}
    \label{Question4p}
  \end{minipage}
   \label{surveyresults}
\end{figure}

Given that there are no previous works addressing the task of behavioural pattern discovery from egocentric photo-streams,  we compare the performance of the proposed pipeline against a native version using a CNN. The native approach uses as images descriptor the softmax output values of a CNN (e.g. Places365 \cite{zhou2017places} model) and the normalized time-stamp. A time-frame is described as the aggregation of the features vectors that compose it. The DBSCAN clustering technique \cite{ester1996density} is used to group similar time-frames. To balance the time characteristic with the rest of the 365 characteristics, we apply a weighting procedure to the input feature vector according to \cite{dousthagh2019feature}. The clusters were grouped into the corresponding time slots and the resulting clusters were considered as patterns. To evaluate quantitatively this patterns and compare them to the ones obtained with our method, we calculated for both approaches, the Silhouette Score for all the users. For our proposed method the Average Silhouette is \textbf{0.12}, with a minimum of -0.13 and a maximum of 0.47. For DBSCAN the average is \textbf{0.01}, with a minimum of -0.49 and a maximum of 0.44. It is important to highlight that the average of clusters found by DBSCAN was 3.0 and in our method was 5.6, even so in this last one there was less overlap.\\ 
Regarding the survey responses, as shown in Figs. \ref{Question2}, \ref{Question3}, \ref{Question4r} and \ref{Question4p} all the individuals answered they found different patterns on the images. Also, regarding if the patterns found by our method represent the user's behaviour, 61.5\% of them answered "yes", 36.3\% answered "maybe", while 2.2\% answered "no". For the question about which method gets better results, 73.6\% answered our method is better, while 26.4\% answered DBSCAN get better results. Finally, 68.1\% answered our method found "correct patterns", 18.7\% answered "Merged Patterns", 9.9\% answered "Sub-Patterns", while 3.3\% answered "No-Patterns". All these values were averaged over all users.

